# YOLOv3 with Spatial Pyramid Pooling for Object Detection with Unmanned Aerial Vehicles

Wahyu Pebrianto, Panca Mudjirahardjo, Sholeh Hadi Pramono, Rahmadwati, Raden Arief Setyawan

Electrical Engineering, Faculty of Engineering, University of Brawijaya
Jl. M.T. Haryono No.167, Kota Malang, 65145, Indonesia

*E-mail: wahyu.pebrianto1@gmail.com*

**Abstract**

Object detection with Unmanned Aerial Vehicles (UAVs) has attracted much attention in the research field of computer vision. However, not easy to accurately detect objects with data obtained from UAVs, which capture images from very high altitudes, making the image dominated by small object sizes, that difficult to detect. Motivated by that challenge, we aim to improve the performance of the one-stage detector YOLOv3 by adding a Spatial Pyramid Pooling (SPP) layer on the end of the backbone darknet-53 to obtain more efficient feature extraction process in object detection tasks with UAVs. We also conducted an evaluation study on different versions of YOLOv3 methods. Includes YOLOv3 with SPP, YOLOv3, and YOLOv3-tiny, which we analyzed with the VisDrone2019-Det dataset. Here we show that YOLOv3 with SPP can get results mAP 0.6% higher than YOLOv3 and 26.6% than YOLOv3-Tiny at 640x640 input scale and is even able to maintain accuracy at different input image scales than other versions of the YOLOv3 method. Those results prove that the addition of SPP layers to YOLOv3 can be an efficient solution for improving the performance of the object detection method with data obtained from UAVs.

## 1. Introduction

In recent years, object detection with Unmanned Aerial Vehicles (UAVs) has attracted much attention in computer vision research and has provided many benefits in various domains. Such as fire smoke detection [1], military [2], urban surveillance [3], and agriculture [4], [5]. However, it is not easy to accurately detect objects with UAVs that capture object images using the camera from a very high followed by a widely geographic one. Most of the current traditional object detection methods are based only on the sliding-window paradigm and handcrafted features. Like Viola-Jones [6], Histogram of Oriented Gradients (HOG) [7], Scale-Invariant Feature Transform (SIFT) [8], [9], Haar [10], [11], which has made significant progress in the research field of object detection. However, this method takes time and effort to achieve the robustness of feature representation and is still vulnerable to failure when handling variations in data obtained from the UAVs. What is urgently needed by object detection systems with UAVs today is an accurate method capable of processing image data end-to-end. Currently, deep learning [12] is one of the solutions to answer these needs.

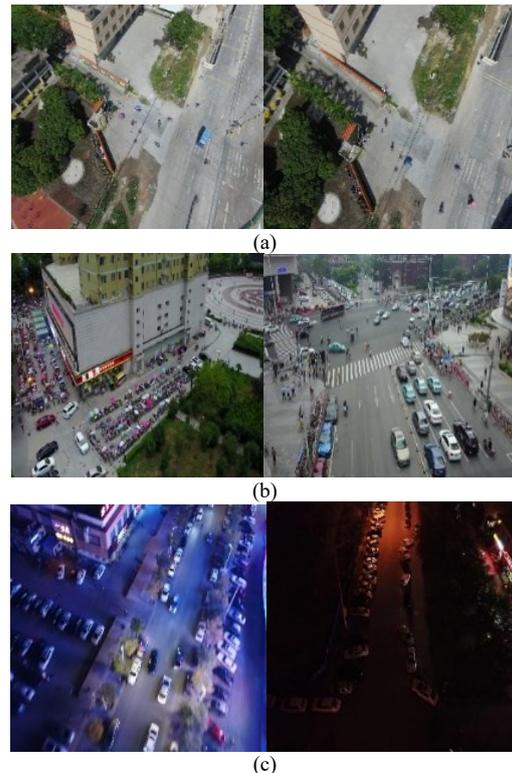

**Fig. 1.** Object detection challenges with UAVs: (a) small objects, (b) object density, and (c) different illuminations[13]

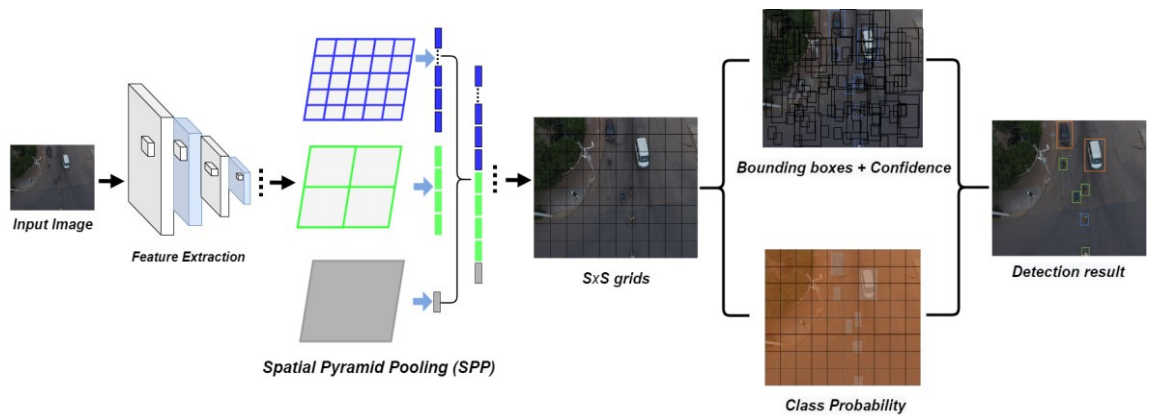

**Fig. 2.** Architecture method in this study

Currently, deep learning methods have become a focus of research in the field of object detection. In particular, deep learning methods based on Convolutional Neural Networks (CNN). The CNN can process visual data accurately without the need to go through a separately feature extraction process and has proven can outperform traditional methods in ImageNet Large Scale Visual Recognition Challenge (ILSVRC) [14]. The progress is inseparable from the availability of large-scale data, such as Microsoft Common Objects in Context (COCO) [15], Pascal Visual Object Classes (PASCAL VOC) [16], ImageNet [14], as well as the availability of computing resources, and driven by ongoing research with the proposed various network architectures. Such as VGG [17], GoogLeNet [18], Residual Networks (ResNets) [19], [20], ResNeXt [21], Cross Stage Partial Network (CSPNet) [22], and EfficientNet [23] in the classification tasks which is widely used as a backbone layer for feature extraction in the object detection tasks. Object detection based on deep learning methods generally divided into two: the one-stage detector and the two-stage detector. The two-stage detector method predicts the bounding box through the process of region proposal and then classifies it to detect the class from the object. Such as the Region-based Convolutional Neural Network (R-CNN) proposed by Ross Girshick *et al.* [24] is the first deep learning based object detection method. R-CNN In the PASCAL VOC 2010 challenge [16] was able to outperform traditional detector methods, such as Deformable Parts Model (DPM) [25], which at that time occupied the first position. This progress is also driven by the development of other popular methods, such as Fast R-CNN [26] and Faster R-CNN [27], which is average have a high prediction accuracy. However, that method is still relatively slow in the detection process. That is deficiency can overcome by one-stage detector methods, such as RetinaNet [28], You Only Look Once (YOLO) [29]–[32] and Single Shot MultiBox Detector (SSD) [33], which are very fast when predicting objects. Such as the YOLO method proposed by Joseph Redmon *et al.* [29] can predict multiple bounding boxes and class probabilities simultaneously, which makes it very fast during the detection process. However, YOLO in the first version still has several localization errors compared with the region proposal method [24], [26], [27]. So development was also carried out to reduce the shortcomings of previous versions, such as YOLOv2 [30] and YOLOv3 [31]. YOLOv2 uses darknet-19 as backbone layers that consist of 19 convolutional layers and 5 max-pooling. While YOLOv3 is a further development of YOLOv2, which can predict the bounding boxes with multi-scale prediction and uses darknet-53 in the backbone layer. YOLOv3 can produce a balance between accuracy and detection speed. The result of YOLOv3 can get the average precision better than Faster R-CNN [27], YOLOv2 [30], SSD [33], and faster than RetinaNet [28] and Region-based Fully Convolutional Network (R-FCN) [34] on the testing of the COCO dataset [15]. However, the data obtained from the UAVs is not like that of data from COCO [15], PASCAL VOC [16], and ImageNet [14], that dominated by global image objects with large individual objects. UAVs capture object images from a very high camera and produce data with varying perspectives viewing. As illustrated in Figure 1, the image data captured by UAVs is dominated by small object sizes, which makes the image contain less clear features and the density of objects with different illumination levels is also a challenge for detecting objects with the images obtained by UAVs.

Motivated by the challenge above, we aim to improve the performance of the YOLOv3 method for detecting objects images obtained from UAVs.

We added Spatial pyramid pooling (SPP) [35] at the end of the darknet-53 backbone architecture to achieve a more efficient feature extraction process. The details objective and the contribution of this study are explained as follows:
1) We improved the performance of YOLOv3 [31] by adding SPP [35] on the end layer of the darknet-53 backbone to obtain more efficient feature extraction process in object detection tasks with UAVs.
2) We also show an evaluation study of different versions YOLOv3 method on object detection tasks with UAVs, including YOLOv3 with SPP, YOLOv3, and YOLOv3-tiny which we analyzed with the VisDrone2019-Det dataset [13].

## 2. Research Method

### 2.1. YOLOv3

You Only Look Once (YOLO) [29] consists of a backbone layer for feature extraction and a head layer for detection. YOLO predicts objects by mapping the image input pixels to $SxS$ grid. Each grid cell predicts $B$ bounding box and confidence score, which is described in the following equation,

$$confidence = P_r(Object) * IoU_{predict}^{truth} \quad (1)$$

$P_r(Object)$ shows the probability of an object inside the bounding box, and $IoU_{predict}^{truth}$ shows the Intersection over Union ($IoU$) of ground truth and box prediction. The confidence will have a value of 0 if there are no objects in the grid cell and a value of 1 if there are objects. The bounding box consists of 5 parameters $(x, y, w, h, confidence)$, the width and height are represented by $w, h$, and $x, y$ represents the center coordinates of the bounding box. In the end, the results of predicted confidence will represent the Intersections over Union ($IoU$) between the predicted box and the ground truth boxes. At the same time, each grid cell also predicts $C$ conditional class probabilities that described in the following equation,

$$Class\ probability = Pr(Class_i|Object) \quad (2)$$

The predicting process of conditional class probabilities $C$ in each grid cell is conditioned if there are objects in the grid cell. And the testing process will multiply of conditional class probability with the predicted value of the box confidence to get the confidence score class specific in each box. As represented by equation (3), which encodes the probability of the class appearing in the box and also represents how well the predicted box matches the object.

$$Pr(Class_i|Object) * P_r(Object) * IoU_{predict}^{truth} = Pr(Class_i) * IoU_{predict}^{truth} \quad (3)$$

YOLOv3 is an improvement over its predecessors [29], [30], which involves different architecture and is more accurate in the detection process. YOLOv3 uses darknet-53 for the feature extraction process, as represented by Figure 3. Darknet-53 uses *3x3* and *1x1* convolutional layers of darknet-19 in YOLOv2 [30], which is organized by residual networks [19]. YOLOv3 predicts bounding boxes with three different scales using ideas from Feature Pyramid Network (FPN) [36], where the final feature map results from the convolutional layers will predict *3D* tensors which are coded as bounding boxes, objectness, and class predictions. Each scales predict 3 squares which are represented as $SxSx(3*(4+1+80))$, where $SxS$ represent the size of the feature map, 4 bounding boxes, 1 objectness prediction, and 80 which is illustrated as the total class prediction.

| | Type | Filters | Size | Output |
|---|---|---|---|---|
| | Convolutional | 32 | 3 × 3 | 256 × 256 |
| | Convolutional | 64 | 3 × 3 / 2 | 128 × 128 |
| 1× | Convolutional | 32 | 1 × 1 | |
| | Convolutional | 64 | 3 × 3 | |
| | Residual | | | 128 × 128 |
| | Convolutional | 128 | 3 × 3 / 2 | 64 × 64 |
| 2× | Convolutional | 64 | 1 × 1 | |
| | Convolutional | 128 | 3 × 3 | |
| | Residual | | | 64 × 64 |
| | Convolutional | 256 | 3 × 3 / 2 | 32 × 32 |
| 8× | Convolutional | 128 | 1 × 1 | |
| | Convolutional | 256 | 3 × 3 | |
| | Residual | | | 32 × 32 |
| | Convolutional | 512 | 3 × 3 / 2 | 16 × 16 |
| 8× | Convolutional | 256 | 1 × 1 | |
| | Convolutional | 512 | 3 × 3 | |
| | Residual | | | 16 × 16 |
| | Convolutional | 1024 | 3 × 3 / 2 | 8 × 8 |
| 4× | Convolutional | 512 | 1 × 1 | |
| | Convolutional | 1024 | 3 × 3 | |
| | Residual | | | 8 × 8 |
| | Avgpool | | Global | |
| | Connected | | 1000 | |
| | Softmax | | | |

**Fig. 3.** Darknet-53 Architecture

### 2.2 Spatial Pyramid Pooling

Spatial Pyramid Pooling (SPP) [37], [38] in CNN was first introduced by [35]. The process of SPP is represented in Figure 4, which receives the input feature map from the convolutional layers. Afterward, in each spatial bin, the pooling layer responds to each filter to produce output *(kM-vector)*. *M* represents the number of bins, $k$ is the number of filters in the last convolutional layer, and the fixed dimensional vector is the input to the fully connected layers. SPP has some extraordinary properties for deep CNN compared with general networks that use pooling sliding windows. Based on research that conducted by He *et al.* [35] SPP-net is capable of producing the output of fixed-length regardless of the input size and uses multi-level spatial bins. Meanwhile, pooling sliding windows only use single window sizes. In this study, we aim to add SPP to the final layer of darknet-53 in YOLOv3 to improve performance in object detection tasks with UAVs.

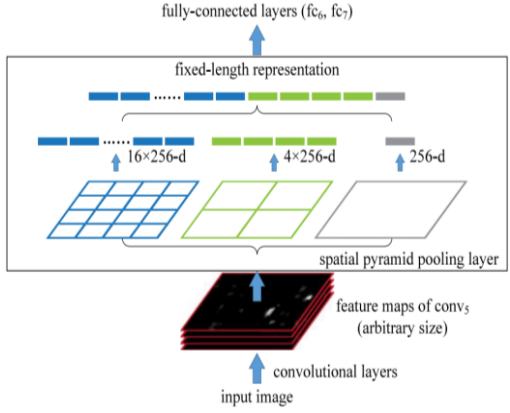

**Fig. 4.** Spatial Pyramid Pooling architecture

### 2.3 Architecture Model

In this study, we aim to add an SPP layer to the final darknet-53 layer to improve the performance of YOLOv3 in object detection tasks with data obtained from UAVs. The details of the architecture in this study is represented in Figure 2. The first process is feature extraction from the input image with darknet-53. Then the SPP layer is added to the final darknet-53 layer to improve the feature extraction process. In the end, the results of SPP is a feature map that uses as input into head detection of YOLOv3 for predicting the bounding boxes and class probabilities.

### 2.4 Loss Function

In Los Function is used to determine the state of the training model in each iteration to calculate the difference between the value of predicted and the value of ground truth. As represented in equation (4), this study split three loss functions: $(l_{coord}, l_{IoU}, l_{class})$. The notation of $l_{coord}$ represents the coordinate prediction errors, $l_{IoU}$ is $IoU$ errors, and $l_{class}$ is the classification errors.

$$Loss = l_{coord} + l_{IoU} + l_{class} \quad (4)$$

The coordinate prediction error is represented in the following equation,

$$l_{coord} = \lambda_{coord} \sum_{i=0}^{S^2} \sum_{j=0}^{B} I_{ij}^{obj}[(x_i - \hat{x}_i)^2 + (y_i - \hat{y}_i)^2] + \lambda_{coord} \sum_{i=0}^{S^2} \sum_{j=0}^{B} I_{ij}^{obj}\left[(\sqrt{w_i} - \sqrt{\widehat{w}_i})^2 + (\sqrt{h_i} - \sqrt{\widehat{h}_i})^2\right] \quad (5)$$

Where $\lambda_{coord}$ is the weight coordinate error, $S^2$ is the number of grid cells for each detection layer, and $B$ is the number of bounding boxes in each grid cell. $(x_i, y_i, \hat{x}_i, \hat{y}_i)$ represents the center coordinate of the ground truth and the target object. Whereas $(h_i, w_i, \hat{h}_i, \widehat{w}_i)$ represents the width and height of the ground truth and the target prediction box. For $IoU$ errors and Classification errors are denoted by equations (6) and (7) as follows,

$$l_{IoU} = \lambda_{IoU} \sum_{i=0}^{S^2} \sum_{j=0}^{B} I_{ij}^{obj}[(C_i - \hat{C}_i)^2]$$

$$+ \lambda_{noobj} \sum_{i=0}^{S^2} \sum_{j=0}^{B} I_{ij}^{noobj}(C_i - \hat{C}_i)^2 \quad (6)$$

$$l_{class} = \lambda_{class} \sum_{i=0}^{S^2} I_i^{obj} \sum_{c \in classes}(p_i(c) - \hat{p}_i(c))^2 \quad (7)$$

The $IoU$ error indicates the degree of overlap between the ground truth and the prediction box. If the anchor box indicates that there is a target located in grid cells $(i, j)$, then the value of $I_{ij}^{obj}$ is 1, and otherwise, the value is 0. The notation of $\lambda_{noobj}$ represents a belief penalty if the prediction box contains no objects, and also misclassification, which represents classification accuracy. Where $p_i(c)$ is the value of true probability and $\hat{p}_i(c)$ is the predicted value of the target.

## 3. Experiment

### 3.1 Dataset

In this study, we used the VisDrone2019-Det dataset [13], which consisted of 10,209 images, 6,471 for training, 548 for validation, and 3,190 for testing. The VisDrone2019-Det dataset consists of ten object categories: pedestrian, person, bicycle, car, van, truck, tricycle, awning-tricycle, bus, and motorcycle. As shown in Figure 5. The VisDrone dataset has several objects with different levels of occlusion in each category, which becomes a challenge to detect objects with UAVs. In this study, we use a training set for the training process and evaluate it with set validation.

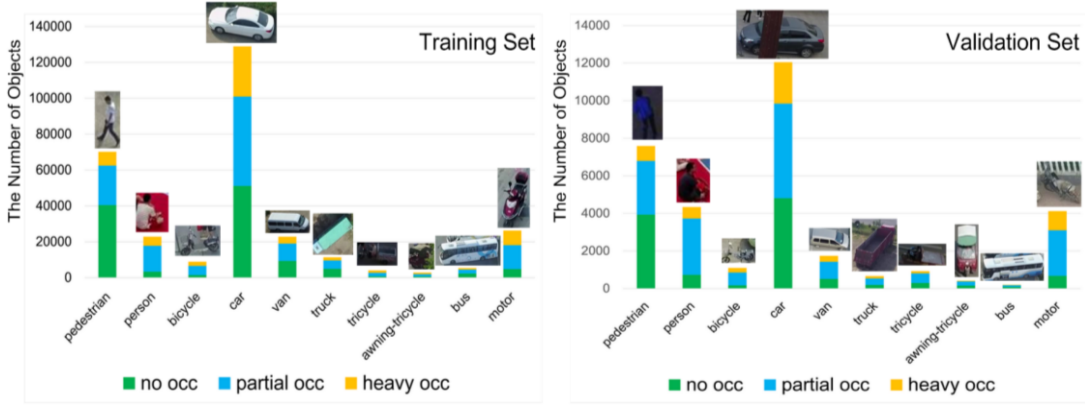

**Fig. 5.** Visdrone dataset with different levels of occlusion

**Table 1.** Training results.

| Model | Precision | Recall | mAP_50 |
|---|---|---|---|
| YOLOv3 | 50.1 | 40.2 | 39.7 |
| YOLOv3-tiny | 22.9 | 17.9 | 13.7 |
| YOLOv3-SPP | 49.3 | 41.4 | 40.3 |

**Table 2.** Detection results.

| Model | Pedestrian | People | Bicycle | Car | Van | Truck | Tricycle | Awn | Bus | Motor |
|---|---|---|---|---|---|---|---|---|---|---|
| YOLOv3 | 49.3 | 39.8 | 16.3 | 78.3 | 41.8 | 38.3 | 25.3 | 12.2 | 49.8 | 45.8 |
| YOLOv3-tiny | 16 | 15.6 | 2.4 | 46.7 | 1 | 10.1 | 6.5 | 2.8 | 10.6 | 16.6 |
| YOLOv3-SPP | 49.4 | 39.4 | 17.8 | 78.1 | 42.8 | 37.9 | 26.7 | 14.3 | 50.6 | 45.6 |

**Table 3.** Validation results with different input scales.

| Model | Input scales | Precision | Recall | mAP_50 |
|---|---|---|---|---|
| YOLOv3 | 960x960 | 48.7 | 40.2 | 38 |
|  | 1280x1280 | 49.4 | 39.8 | 38.2 |
| YOLOv3-tiny | 960x960 | 25.8 | 21.1 | 15.6 |
|  | 1280x1280 | 25.6 | 21.9 | 16.1 |
| YOLOv3-SPP | 960x960 | 48.3 | 41.3 | 38.8 |
|  | 1280x1280 | 47.2 | 42.8 | 39.1 |

### 3.2 Metric Evaluation

To evaluate each method, we used the parameters Precision ($P$), Recall ($R$), Average Precision ($AP$), and mean Average Precision ($mAP$) with 0.5 Intersections over Union ($IoU$). The details of $P$ and $R$ parameters are described by the following equation,

$$Precision\ (P) = \frac{TP}{TP+FP} \quad (8)$$

$$Recall\ (R) = \frac{TP}{TP+FN} \quad (9)$$

Where $TP$ is true positive, that is the correct detection of the ground truth bounding box, $FP$ is false positive, that is object was detected but misplaced. $FN$ is false negative, which means that the basic ground truth of the bounding box was not detected. $AP$ and $mAP$ parameters are described by the following equation,

$$AP = \sum_n (R_{n+1} - R_n) \max_{\tilde{R}:\tilde{R} \geq R_{n+1}} P(\tilde{R}) \quad (10)$$

$$mAP = \frac{1}{N}\sum_{i=1}^{N} AP_i \quad (11)$$

Where $AP$ is the average value of $P$ and $R$. $mAP$ is the average of the $AP$ used to measure all class categories in the dataset and is the metric used to measure the accuracy of object detection with UAV.

### 3.3 Experimental Details

For the experimental procedures in this study, we use a pre-trained model from the COCO dataset [15]. In the training phase, we use stochastic gradient descent as optimization with momentum 0.9, batch size 16, learning rate 0.01, and training iterations of 50 epochs with an input scale 640x640. The framework used in this study is PyTorch with a Tesla T4 GPU for the training and validation process.

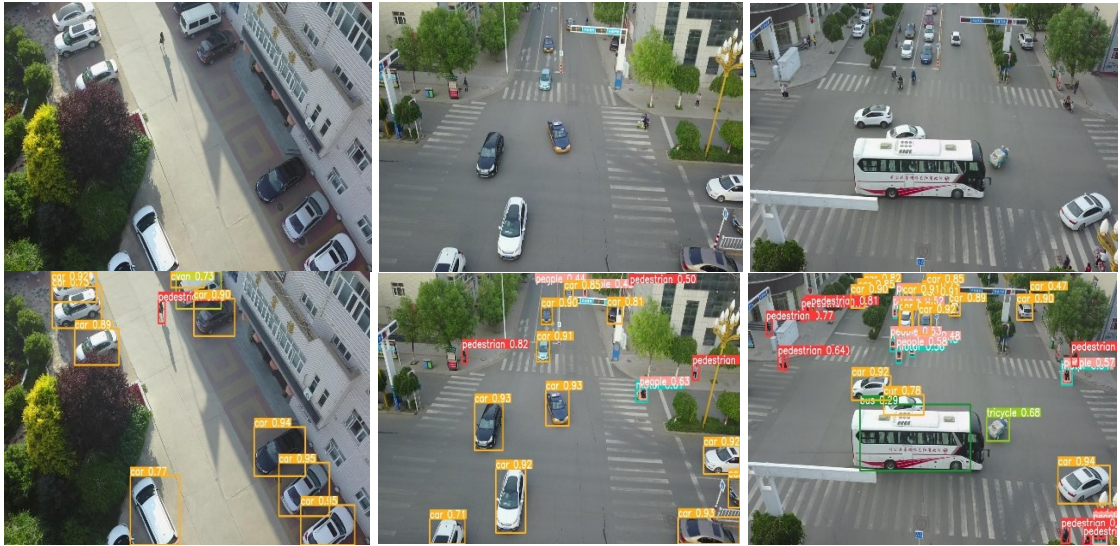

**Fig. 6.** Detection visualization.

## 3.4 Results and Discussion

Table 1, shows the results of training with an input scale of 640x640, which can be concluded in several findings. First, the YOLOv3 with SPP obtained a higher *mAP* of 0.6% than YOLOv3. These results prove that the addition of the SPP architecture to the YOLOv3 can improve the performance of the object detection model. Second, the YOLOv3-tiny obtained the *mAP* value of 26.6% much lower than YOLOv3 with SPP and 26% from YOLOv3. These results, one of which is influenced by the depth of the network. YOLOv3-tiny is a lightweight model with fewer parameters and depth. So that able to obtain faster detection processing. However, inversely proportional to the obtained accuracy. The details of the detection results in Table 1, are shown in Table 2. When observed from the total of 10 detection classes, YOLOv3 with SPP excels in six classes: pedestrian, bicycle, van, tricycle, awn, and bus compared to YOLOv3, which only excels in four classes: people, cars, trucks, and motorcycles. Whereas the results of YOLOv3-tiny is lower than YOLOv3 with SPP and YOLOv3 from all detections. For one of the results of visualization from the detection is represented in Figure 6..

To obtain a more in-depth analysis, we also validate each model with different input scales. Our goal is to find out if the image scale also affects each object detection model. As reported in Table 3. The YOLOv3 with SPP that we propose is still superior to YOLOv3, with an *mAP* difference of 0.8% on a 960x960 scale, and 0.9% on a 1280x1280 scale. Whereas YOLOv3-tiny is still lower on both scales compared to the results of YOLOv3 with SPP and YOLOv3.

## 4. Conclusion

This study aims to improve the performance of YOLOv3 in object detection tasks with UAVs by adding an SPP layer at the end of the darknet-53. We trained three different models: YOLOv3 with SPP, YOLOv3, and YOLOv3-tiny with Visdrone2019-Det training set and evaluated them with a validation set at an input scale of 640x640. The results of YOLOv3 with SPP can improve the performance object detection model, with the results of *mAP* accuracy of 0.6% more height than YOLOv3 and 26.6% than YOLOv3-tiny. The YOLOv3 with SPP also can maintain accuracy at different input scales, which can outperform the results of YOLOv3 with a difference of 0.8% *mAP* accuracy on a 960x960 input scale and 0.9% on a 1280x1280 scale. Meanwhile, YOLOv3-tiny is still lower on both scales compared to the results of YOLOv3 with SPP and YOLOv3. The results of YOLOv3 with SPP prove that the addition of the SPP layers to YOLOv3 can improve the performance of object detection models with data obtained from UAVs even with different input scales of image.